\documentclass[runningheads,a4paper]{llncs}

\usepackage{times}
\usepackage{amssymb}
\usepackage{graphicx}
\usepackage{url}
\usepackage[lined,plain,commentsnumbered,linesnumbered]{algorithm2e}

\urldef{\mailsa}\path|thiago.castro.ferreira@gmail.com, ivandre@usp.br|

\begin{document}

\title{Trainable Referring Expression Generation using Overspecification Preferences}
\titlerunning{Trainable REG using Overspecification Preferences}
\author{Thiago Castro Ferreira \and Ivandr\'e Paraboni}
\authorrunning{T.C. Ferreira \and  I. Paraboni}
\institute{Tilburg University, Tilburg center for Cognition and Communication, Tilburg, Netherlands \and  University of S\~ao Paulo, Schools of Arts, Sciences and Humanites, S\~ao Paulo, Brazil 
\mailsa\\
}

\index{Ferreira, Thiago}
\index{Paraboni, Ivandr\'e}

\toctitle{} \tocauthor{}

\maketitle

\begin{abstract}
Referring expression generation (REG) models that use speaker-dependent information require a considerable amount of training data produced by every individual speaker, or may otherwise perform poorly. In this work we present a simple REG experiment that allows the use of larger training data sets by grouping speakers according to  their overspecification preferences. Intrinsic evaluation shows that this method generally outperforms the personalised method found in previous work.
\end{abstract}

\section{Introduction}

In natural language generation systems, referring expression generation (REG) is the microplanning task responsible for generating descriptions of discourse objects \cite{survey}. REG includes the well-known content selection task for definite description generation, which is the focus of the present work.

Existing work in computational REG  and related fields have identified a wide range of factors that may drive content selection. To a considerable extent, however, content selection is known to be influenced by human variation \cite{viethen-speaker-dep}. In other words, under identical circumstances (i.e., in the same referential context), different speakers will often produce different descriptions. 

Differences across speakers may be observed in at least two aspects of referential behaviour: (a) in the choice of attributes (e.g., `the large ball' vs, `the red ball') and (b) in the level of referential overspecification (e.g., `the ball' vs. `the {\em red} ball' in a context in which there is only one ball.) In this work we will focus on the issue of overspecification (b), discussing how preferences of this kind may be taken into account in trainable, speaker-dependent REG.

Existing REG algorithms as in \cite{Bohnet2008,nle} usually pay regard to human variation by computing personalised features from a training set of descriptions produced by each individual speaker. This highly personalised training method may of course be considered an ideal account of human variation but, in practice, will only be effective if every speaker in the domain is represented by a sufficiently large number of training instances.  

As an alternative to standard speaker-dependent REG, in this work we describe a simple experiment in which a machine-learning REG model is trained on descriptions produced by a group of speakers with similar referential behaviour (i.e., as opposed to using only the descriptions produced by each speaker individually.)  By allowing the use of larger training datasets in this way, we would like to show that this method may  outperform the use of personalised information addressed in previous work, an improvement that may be particularly useful when the availability of training examples produced by every speaker is limited.

\section{Related Work}
\label{sec:related}

Existing methods for speaker-dependent REG generally consist of computing the relevant features for each individual speaker. In what follows we summarise a number of studies that follow this method - which is equivalent to our {\em Speaker} baseline method to be discussed in Section \ref{sec-training-methods}.

In \cite{Bohnet2008}, the Incremental algorithm \cite{incremental} and a number of extensions of the Full Brevity algorithm \cite{greedy} are evaluated on TUNA data \cite{tuna}. In the case of the Incremental algorithm, human variation is accounted for by computing individual preference lists based on the attribute frequency of each individual speaker as observed in the training data. In the case of Full Brevity, all possible descriptions for a given referent are computed, and the description that most closely resembles those produced by the speaker is selected using a nearest neighbour approach.

The work in \cite{difabbrizio} also makes use of the Full Brevity \cite{greedy} and Incremental \cite{incremental} algorithms to generate TUNA descriptions. In the case of Full Brevity, human variation is accounted for by selecting the set of attributes computed by the algorithm according to the frequency and recency estimate use for each individual speaker. In the case of the Incremental algorithm, human variation is implemented as in \cite{Bohnet2008}, that is, by computing individual preference lists for each speaker.

The work in \cite{viethen-speaker-dep} makes use of decision-tree induction to predict content patterns (i.e., full attribute sets representing actual referring expressions) from GRE3D3/7 data \cite{gre3d3,gre3d7}. Human variation is accounted for by modelling speaker identifiers as machine learning features. 

Finally, the work in \cite{thiago-svm,thiago-speaker-pref,nle} presents a SVM-based approach to speaker-dependent REG tested on GRE3D3/7  and Stars/Stars2 \cite{stars,stars2} data. Once again, human variation is accounted for by computing individual preference lists from the subset of descriptions produced by each individual speaker.  As this  approach  will be taken as the basis of our current work, further details will be discussed in the next section.

\section{Experimental Setup}
\label{sec:experiment}

We designed an experiment to compare two training methods for speaker-dependent REG. Both methods are based on the REG model described in Section \ref{sec-basic-model}. The methods themselves are described in Section \ref{sec-training-methods}.

\subsection{Basic REG model}
\label{sec-basic-model}

Our experiment makes use of a speaker-dependent REG model adapted from \cite{nle}. Given a set $D$ of domain objects, a set $A$  of referential attributes, a set $R$ of spatial relations between object pairs, and a target object $t \in D$  to be identified, content selection is implemented with the aid of a set of classifiers $C_{atom} =\{c^{(1)}, c^{(2)}, ..., c^{(|A|)} \}$, in which $c^{(i)} \in C_{atom}$ predicts whether $a^{(i)} \in A$ should be selected or not, and a multi-class classifier $C_{rel}$ predicts the kind of relation ($r \in R$) that may hold between the target $t$ and the nearest landmark $lm$. $R$  includes the special \textit{no-relation} property to denote situations in which no relation between a certain object pair is predicted. When a relation to a landmark object $lm$ exists, we also consider a set of classifiers $C_{atom}^{lm} =\{c^{(1)}, c^{(2)}, ..., c^{(|A|)} \}$ to describe $lm$.

Part of the input to the classifiers consists of feature vectors extracted from the referential context. These features - hereby called context features - are based on the ones proposed in \cite{viethen-speaker-dep}, and are intended to model target and landmark properties (if any), and similarities between objects. More specifically, context features represent the size of the target and its nearest landmark, the relation between the two objects (horizontal or vertical) and the number of distractors that share a certain property (e.g., {\em type}, {\em colour} etc.) with each of them. 

In order to model human variation, we also consider two kinds of speaker-dependent feature: those that  model personal information about the speakers, and those that model their content selection preferences. Speaker's personal features consist of a unique speaker identifier as in \cite{viethen-speaker-dep}, gender and age bracket. Speaker's preferences consist of lists of preferred attributes for reference to target and landmark objects sorted by frequency.

Attributes and relations of the main target $t$ and nearby landmark $lm$ are combined to form a description $L$ according to Algorithm \ref{alg:combining}.

\begin{algorithm}[htp]
\caption{Classification-based REG}
\label{alg:combining}
\footnotesize{
	\DontPrintSemicolon
	\SetKwFunction{algo}{getDescription}
	\SetKwProg{myalg}{Algorithm}{}{}
	\myalg{\algo{$t$, $L$, $D$, $H$}}{
	$L[t] \leftarrow \{\}$ \;
	$H \leftarrow H \cup t$ \; 
	$level \leftarrow |H|$ \;
	$Pr_{atom} \leftarrow getPredictions(level)$ \;
	$Pr_{rel} \leftarrow getRelationPrediction(level)$ \;
	
	\For{ $A_{i} \in Pr_{atom}$}{
		\If {$Pr_{atom}[A_{i}]==1$} {
 			$L[t] \leftarrow L[t] \cup \langle A_{i}, value(t,A_{i}) \rangle$ \;
		}
	}
	
	\If {$Pr_{rel} \neq no$-$relation$} {
		$lm \leftarrow value(t, Pr_{rel})$\;
		\If{$lm \neq null$ \textbf{and} $lm \notin H$} {
			$L[t] \leftarrow L[t] \cup \langle rel, lm \rangle$ \;
			$L \leftarrow getDescription(lm, L, D, H)$ \;
		}
	}	
	\KwRet $L$\;}{}
	}
\end{algorithm}

The input to the algorithm is a target $t$ and a domain $D$. The algorithm also makes use of a history list $H$ to prevent self-reference (e.g., `the ball next to a box that is next to a ball that...') and the initially empty  list $L$ representing the output description (to be built recursively). 

An auxiliary function $level$ is assumed to return 1 when $t$ corresponds to the main target, 2 when $t$ corresponds to the first landmark object, and so on. This information is taken into account to invoke the appropriate set of classifiers, which are implemented by the auxiliary functions $getPredictions$ and $getRelationPrediction$. The former is assumed to invoke the set of binary classifiers for every attribute of $t$, and the latter invokes the multivalue prediction for the $relation$ class. 

Content selection proper is performed by selecting all atomic attributes of the target $t$ that were predicted by the corresponding binary classifiers.  If a relation between $t$ and its nearest distractor $lm$ has been predicted, the relation is included in $L$ and the algorithm is called recursively to describe $lm$ as well. As in \cite{nle}, all classifiers are built using Support Vector Machines (SVMs) with a Gaussian Kernel. For the relation prediction, we use an ``one-against-one'' multi-class method.

\subsection{Training methods}
\label{sec-training-methods}

We consider two training methods for the basic REG model described in the previous section: a baseline method called  \textit{Speaker}, and our proposed \textit{Profile} method. 

In the \textit{Speaker} method, classifiers are trained on the set of referring expressions produced by each individual speaker as in \cite{Bohnet2008}. This method will effectively create personalised REG models, and may in principle be considered ideal for the purpose of modelling human variation in REG. In order to be successful, however, the method requires a sufficiently large number of descriptions produced by every single speaker, which may not always be available.

As an alternative to the standard {\em Speaker} approach, we propose a  training method based on the simple observation - made by \cite{viethen-speaker-dep} and others - that some speakers follow a consistent pattern in reference production, whereas others do not. More specifically, in the present method - hereby called {\em Profile} - speakers are divided into three general categories: those that always produced overspecified descriptions, those that always produced minimally distinguishing descriptions, and those that do not follow a consistent pattern. Knowing in advance the category of a particular speaker, the REG model will be trained on the subset of descriptions produced by that category only. This will effectively allow us to use more training data than in the {\em Speaker} method, and it should  improve the overall results of the REG model.

\subsection{Evaluation}
\label{sec-eval}

The {\em Speaker} and {\em Profile} training methods were compared against each other using six REG datasets: TUNA-Furniture and TUNA-People - only descriptions to single objects were considered -, GRE3D3, GRE3D7, {Stars} \cite{stars} and {Stars2} \cite{stars2}.

All models were built using cross-validation with a balanced number of referring expressions per participant within each fold.  For TUNA and {Stars}, descriptions were divided into six folds each. For GRE3D3/7 and Stars2, descriptions were divided into ten folds each. 


Optimal values for the SVM $C$  parameter and for the Gaussian kernel $\gamma$ parameter were obtained using grid-search. We tested $C$ values of 1, 10, 100 and 1000 and $gamma$ values of 1, 0.1, 0.01, and 0.001 in a validation set before testing the models. Given $k$ folds in a cross-validation iteration, $k-2$ folds were used as training data, one fold was used to estimate the optimal values of $C$ and $\gamma$, and the remaining fold was used to test the model. 

We measured Dice coefficients \cite{dice} to assess the similarity between each description generated by the model and the corpus description. We also computed the overall REG Accuracy by counting the number of exact matches between each description pair.

\section{Results}

Table \ref{table:results} presents the results of the REG model using the {\em Speaker} and {\em Profile} training methods on each of the test domains.

\begin{table*}[h]
\begin{center}
\begin{tabular}{l | c  c    c c  |  c c     c c  |   c c     c c	| c c}
\hline
\multicolumn{1}{c|}{} 
& \multicolumn{2}{c}{TUNA-f}
& \multicolumn{2}{c|}{TUNA-p}
& \multicolumn{2}{c}{GRE3D3}
& \multicolumn{2}{c|}{GRE3D7}
& \multicolumn{2}{c}{Stars}
& \multicolumn{2}{c|}{Stars2}
& \multicolumn{2}{c}{Overall}
\\
Method							&Dice	&	Acc.			&	Dice								&	Acc.			&	Dice								&	Acc.		&	Dice&	Acc.&	Dice&	Acc.&	Dice&	Acc.&	Dice&	Acc.\\
\hline
{Speaker} & 0.85 & 0.41 & 0.71 & 0.24 	& 0.88 & 0.61 	& 0.92 & 0.72 & 0.75 & \textbf{0.39} & 0.70 & 0.31 & 0.87 & 0.60 \\
{Profile} & 0.85 & 0.43 & \textbf{0.78}	& \textbf{0.35} 		& \textbf{0.93} 	& \textbf{0.74} & \textbf{0.94} & \textbf{0.77} & 0.73 & 0.32 & \textbf{0.78} & \textbf{0.40}  & {\bf 0.90} & \textbf{0.66} \\
\hline
\end{tabular}
\caption{Content selection results}
\label{table:results}
\end{center}
\end{table*}

Overall results  suggest that the {\em Profile} training method outperforms {\em Speaker} in terms of Dice (Wilcoxon W$=$3188296.5, p$<$.01) and Accuracy (Chi-Square $\chi^{2}=$104.28, p$<$.01). The main exception is the Stars corpus, in which the {\em Profile} model failed to accurately predict the use of relational properties that are ubiquitous in this domain. More work will be required to shed light on this particular issue.

Results based on Dice coefficients were also confirmed in four individual domains: TUNA-People (W$=$17969, p$<$.01), GRE3D3 (W$=$21483, p$<$.01), GRE3D7 (W$=$759954, p$<$.01) and Stars2 (W$=$100727, p$<$.01). In the case of TUNA-Furniture and Stars the difference between the two methods was not significant.

Regarding Accuracy, results were also confirmed in four individual domains. TUNA-People ($\chi^{2}=$19.61, p$<$.01), GRE3D3 ($\chi^{2}=$ 61.71, p$<$0.01), GRE3D7 ($\chi^{2}=$64.61,p$<$0.01) and Stars2 ($\chi^{2}=$27.97,p$<$0.01). In the case of TUNA-Furniture the difference between the two methods was not significant, and in the case of Stars a significant effect in the opposite direction was observed ($\chi^{2}=$9.38, p$<$0.01).

Given that speakers are grouped according to their overspecification preferences, it is interesting to observe whether our output descriptions actually correspond to the expected level of information. To this end, Table \ref{table:results2} shows  how often the {\em Speaker} and {\em Profile} methods were able to reproduce the level of referential specification found in each corpus, that is, how often each method correctly produced underspecified, overspecified and minimally distinguishing descriptions. Results show that predictions made by the {\em Profile} method generally outperform those made by the {\em Speaker} method. The exception  is, once again the Stars domain as discussed above.

\begin{table*}[h]
\begin{center}
\begin{tabular}{l |  c  c  |  c     c  |   c     c	|   c}
\hline
Method							&TUNA-f		& TUNA-p	& GRE3D3		&	GRE3D7	&	Stars		&		Stars2	&	Overall\\
\hline
{Speaker} 					& 0.75 			& 0.70				& 0.54					& 0.80				& 0.70			& 0.65				& 0.75\\
{Profile} 						&0.78				& 0.78				& 0.61					& 0.82				& 0.68			& 0.78				& 0.79
\\
\hline
\end{tabular}
\caption{Referential overspecification accuracy}
\label{table:results2}
\end{center}
\end{table*}

Finally, Table \ref{table:class-results}  shows Precision, Recall and F1-measures obtained by both methods according to reference type (minimally distinguishing, overspecified and underspecified). Results show that both models generally make accurate predictions regarding the generation of overspecified descriptions, which make the majority of our data. For underspecified and minimally distinguishing descriptions, on the other hand, results remain much lower due to data sparsity.

\begin{table}[h]
\begin{center}
\begin{tabular}{l  c     |    c  c  c    | c   c  c}
\hline
\multicolumn{1}{c}{} 
&\multicolumn{1}{c|}{} 
&\multicolumn{3}{c|}{Speaker}
&\multicolumn{3}{c}{Profile}\\
Reference type&			support		&		P					& 		R			& 		F &	P				& 		R			& 		F 		\\
\hline
{Minimal.}& 		1219				&		0.68      	& 0.21     & 0.32  &	0.75     &  0.22  	&  0.34    \\
{Oversp.} 	&			5777				&		0.85     	& 0.86     & 0.86  &	0.85     & 0.91    	&  0.88    \\
{Undersp.}&			162					&		0.11      	& 0.61     & 0.19  &	0.15     &  0.56    &  0.24     \\
{Overall}						&			7158				&		0.80      	& 0.75     & 0.75  &	0.82     & 0.79   	&  0.77    \\
\hline
\end{tabular}
\caption{Reference type classification results }
\label{table:class-results}
\end{center}
\end{table}

\section{Final remarks}

This paper presented an experiment in machine-learning REG that takes speaker-dependent information into account, and which makes use of a  simple training method based on speaker profiles to circumnavigate the issue of data sparsity. By grouping speakers according to their overspecification preferences, we were able to sketch a speaker-dependent REG model that was shown to outperform the standard use of individual speaker's information proposed in previous work.

Despite the overall positive results of this initial experiment, we may of course ask which alternative training methods may be considered for the task. More specifically, since using more training data - as we did by considering groups of similar speakers -  has improved results, it may be the case that by simply training our REG models on the data provided by {\em all} speakers, we could improve results even further. Although we presently do not seek to validate this claim (which in any case would defeat the purpose of using speaker-dependent information in REG), there is plenty of evidence to suggest that this would not be the case. Studies such as in \cite{Bohnet2008,Viethen2010}, for instance, have consistently shown that using individual training datasets for each speaker outperforms speaker-independent REG and, in particular, the work in \cite{nle} has shown that SVM-based REG models generally produce best results when trained on personalised datasets.

Finally, we notice that the present experiment has focused on a single aspect of referential behaviour, namely, on the issue of overspecification preferences across speakers. As future work, we would like not only to refine the current method (e.g., by distinguishing between target and landmark overspecification preferences, among many other options.) but also to consider the issue of attribute choice (e.g., by grouping speakers according to their preferred referential attributes.)

\section*{Acknowledgements}
This work has been supported by the National Council of Scientific and Technological Development from Brazil (CNPq) and FAPESP.


\bibliography{main}
\bibliographystyle{splncs}

\end{document}